# Online and Offline Handwritten Chinese Character Recognition: A Comprehensive Study and New Benchmark

Xu-Yao Zhang[1], Yoshua Bengio[2], Cheng-Lin Liu[1]

1: NLPR, Institute of Automation, Chinese Academy of Sciences, P.R. China
2: MILA, University of Montreal, Canada

**Abstract**

Recent deep learning based methods have achieved the state-of-the-art performance for handwritten Chinese character recognition (HCCR) by learning discriminative representations directly from raw data. Nevertheless, we believe that the long-and-well investigated domain-specific knowledge should still help to boost the performance of HCCR. By integrating the traditional normalization-cooperated direction-decomposed feature map (directMap) with the deep convolutional neural network (convNet), we are able to obtain new highest accuracies for both online and offline HCCR on the ICDAR-2013 competition database. With this new framework, we can eliminate the needs for data augmentation and model ensemble, which are widely used in other systems to achieve their best results. This makes our framework to be efficient and effective for both training and testing. Furthermore, although directMap+convNet can achieve the best results and surpass human-level performance, we show that writer adaptation in this case is still effective. A new adaptation layer is proposed to reduce the mismatch between training and test data on a particular source layer. The adaptation process can be efficiently and effectively implemented in an unsupervised manner. By adding the adaptation layer into the pre-trained convNet, it can adapt to the new handwriting styles of particular writers, and the recognition accuracy can be further improved consistently and significantly. This paper gives an overview and comparison of recent deep learning based approaches for HCCR, and also sets new benchmarks for both online and offline HCCR.

## 1. Introduction

Handwritten Chinese character recognition (HCCR) has been studied for more than fifty years [1, 2] to deal with the challenges of large number of character classes, confusion between similar characters, and distinct handwriting styles across individuals. According to the type of input data, handwriting recognition can be divided into online and offline. In online HCCR, the trajectories of pen tip movements are recorded and analyzed to identify the linguistic information expressed [3], while in offline HCCR, character (gray-scaled or binary) images are analyzed and classified into different classes. Offline HCCR finds many applications, such as mail sorting [4], bank check reading, book and handwritten notes transcription, while online HCCR has been widely used for pen input devices, personal digital assistants, smart phones, computer-aided education, and so on. Moreover, HCCR is also an important integral part for handwritten text recognition (both online [5] and offline [6]) which considers segmentation and recognition simultaneously. High character recognition accuracy is essential for the success of handwritten text/string recognition [7].

To promote academic research and benchmark on HCCR, the National Laboratory of Pattern Recognition (NLPR), Institute of Automation of Chinese Academy of Science (CASIA), has organized three competitions at CCPR-2010 [8], ICDAR-2011 [9], and ICDAR-2013 [10]. The results of competition show improvements over time and involve many different recognition methods. An overwhelming trend is that deep learning based methods gradually dominate the competition. From the very beginning, all submitted systems at CCPR-2010 were traditional methods. In ICDAR-2011, the team of IDSIA from Switzerland submitted their system [11] based on convolutional neural network (convNet) and won the first place on offline HCCR. This is the first work on using convNet for HCCR. Later for ICDAR-2013, both the winners of online and offline HCCR were using convNets. The team from Fujitsu R&D Center used a 4-convNet voting method to win the competition of offline recognition, while the team from University of Warwick used a sparse convNet [12] to win the competition of online recognition.

Deep learning methods can directly learn discriminative representations [13] from raw data, and therefore can provide end-to-end solutions for many pattern recognition problems. However, the well-studied domain-specific knowledge is shown to be still helpful for further improving the performance [14, 15] of HCCR. The most important domain knowledge of HCCR includes the character shape normalization and direction-decomposed feature maps. The character recognition community has proposed many useful shape normalization methods such as nonlinear normalization [16], bi-moment normalization [17], pseudo 2D normalization and line density projection interpolation [18]. Shape normalization can reduce the within-class variations and hence increase the recognition accuracy [19]. Another important domain knowledge is the direction-decomposed feature map. By decomposing the gradient (for offline image) or the local stroke (for online stroke trajectory) into different directions (from $0°$ to $360°$), we can obtain multiple feature maps, each representing a direction of original gradient/stroke. This is a strong prior knowledge of Chinese character which is produced by basic directional strokes during writing process. Representing Chinese character as directional features had been the state-of-the-art method [19, 20, 21] for a long time before the arrival of convNet.

To improve the accuracies of HCCR, instead of training convNet from raw data, we represent both the online and offline handwritten characters by the normalization-cooperated [22] direction-decomposed feature maps (directMap), which can be viewed as a $d \times n \times n$ sparse tensor ($d$ is the number of quantized directions and $n$ is the size of the map). DirectMap contains the domain-specific knowledge of shape normalization and direction decomposition, and hence is a powerful representation for HCCR. Furthermore, inspired by the recent success of using very deep convNet for image classification [23, 24, 25], we developed an 11-layer convNet for HCCR. By combining directMap with convNet, we are able to obtain new benchmarks for both online and offline HCCR on the ICDAR-2013 competition database [10]. Previous works usually adopt different methods to obtain best performance for online and offline HCCR separately. However, with directMap+convNet, we are able to achieve state-of-the-art performance for both online and offline HCCR under the same framework. Due to the embedded domain-specific knowledge, we can also eliminate the needs of data augmentation and model ensemble, which are crucial for other systems to achieve their best performance. This makes our model to be efficient and effective for both the training and testing processes.

The large variability of handwriting styles across individuals is another challenge for HCCR. Writer adaptation [26, 27] is widely used to handle this challenge by gradually reducing the mismatch between writer-independent system and particular individuals. Although deep learning based methods have set a high record for HCCR which already surpass human-level performance, we show that writer adaptation in this case is still effective. Inspired from our early work on style transfer mapping [28], we add a special adaptation layer in the convNet to match and eliminate the distribution shift between training and test data in an unsupervised manner. The adaptation can guarantee performance improvements even when only a small number of samples are available, due to the regularization involved in the learning process. During our experiments on 60 writers for both online and offline HCCR, we observed consistent and significant increase of accuracies by the adaptation of the convNet.

The handwriting recognition community has reported many useful and important achievements (from the year of 1980 to 2008) by previous overview papers of [29, 30, 31, 32, 3, 33]. Nowadays, the deep learning based approaches become the new cutting-edge technology for solving handwriting related problems. This paper can be viewed as an overview of recent progresses (especially through the three competitions [8, 9, 10]) in using deep learning methods for the task of handwritten Chinese character recognition (HCCR). The results and comparisons reported here can be used as new benchmarks for future researches in the field of both online and offline HCCR.

The rest of this paper is organized as follows. Section 2 reviews related works. Section 3 describes the procedures for generating online and offline directMaps. Section 4 shows the evolution from traditional methods to convNet. Section 5 introduces the details of the convNet used in our system. Section 6 explains how to add an adaptation layer in convNet for writer adaptation. Section 7 reports the experimental results, and Section 8 draws concluding remarks.

## 2. Related Works

With the impact from the success of deep learning [34, 35] in different domains, the solution for HCCR has been changed from traditional methods to convolutional neural networks (convNet) [36]. The first reported successful use of convNet for HCCR (offline) was the multi-column deep neural network (MCDNN) [37, 38]. After that, the sparse convNet [39] was used to achieve the best performance for online HCCR in ICDAR-2013 competition. Alternately trained relaxation convolutional neural network was proposed by [40] for offline HCCR. Recently, the highest accuracy for offline HCCR was achieved by [41] through integrating multiple strategies such as local and global distortions, multi-supervised training, and multi-model voting. ConvNet has also been successfully used for handwritten Hangul recognition [42] which is similar to HCCR. Although these methods have outperformed traditional methods by large margins, they are based on end-to-end learning which ignores the long-and-well studied domain-specific knowledge in HCCR.

Recently, [15] combined the traditional feature extraction methods such as Gabor and gradient feature maps with the GoogLeNet [24] to obtain very high accuracy for offline HCCR. Moreover, for online HCCR, [14] and [43] achieved the best performance by using convNet with various domain knowledge including deformation, imaginary stroke map, path signature map, and directional map. These results clearly identify the advantages of using domain knowledge for further improving performance. It should be noted that in the application of deep learning to most image classification tasks, the generation of distorted images for augmenting the training data is also a kind of utilization of domain knowledge. However, in our mind, the most important domain-specific knowledge should be shape normalization and direction decomposition. With our proposed directMap+convNet, we can achieve new benchmarks for both online and offline HCCR, without the help from data augmentation or model ensemble, which are crucial for [15] and [43] to obtain their best results.

Deep learning based methods have also found applications in other handwriting related problems, such as writer identification [44], hybrid model [45], confidence analysis [46], handwritten legal amounts recognition [47], and text spotting [48]. The convNet can also be combined with the hidden Markov model (HMM) for online handwriting recognition [49]. Recently, the recurrent neural network (RNN) with long-short term memory (LSTM) [50] has been successfully used for handwritten Chinese text recognition without explicit segmentation of characters [51]. The combination of RNN and convNet has also been used for scene text reading by [52] and [53]. It is evident that more and more character recognition related problems will turn their attention to deep learning methods for high performance solutions.

Writer adaptation has been widely used in personalized handwriting recognition systems [26, 27]. Our previous work [28] proposed a framework of style transfer mapping (STM) for the adaptation of different classifiers, which has been further studied by [54, 55, 56]. Previous writer adaptation is mainly focused on traditional classifiers such as the nearest prototype classifier [57] and modified quadratic discriminant function [1]. However, it is still unclear for the writer adaptation of the deep convNet. Traditional approach for adaptation of deep network [58] is to retrain a classification layer that takes the activations of one of the existing network as input features (such as DeCAF [59]). When labeled data are unavailable for the target domain, subspace alignment (embedding) [58] is widely used to minimize the domain shift. In this work, by viewing STM as a new special layer, we can adapt convNet to new styles of particular writers in an unsupervised manner with only a small amount of writer-specific data. The proposed adaptation layer is a simple and basic component for neural networks, and therefore can be easily integrated with different network architectures.

## 3. Direction Decomposed Feature Map

Shape normalization and direction decomposition are powerful domain knowledge in HCCR. Shape normalization can be viewed as a coordinate mapping in continuous 2D space between original and normalized characters. Therefore, direction decomposition can be implemented either on original (normalization-cooperated) or normalized (normalization-based) characters [22]. The normalization-cooperated method maps the direction elements of original character to directional maps without generating normalized character, and thus can alleviate the effect of stroke direction distortion caused by shape normalization and provide higher recognition accuracies [22]. We use normalization-cooperated method to generate directMaps for both online and offline HCCR [19].

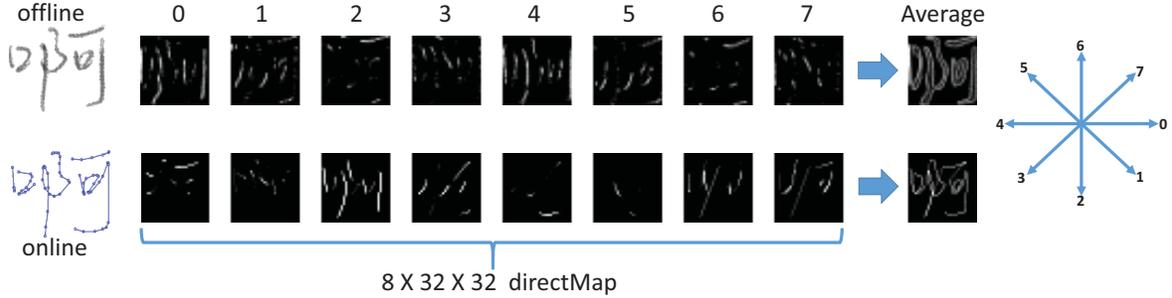

Figure 1: The directMaps for online and offline handwritten Chinese characters.

*3.1. Offline DirectMap*

The offline HCCR datasets provide gray-scaled images with background pixels labeled as 255. For the purpose of fast computation, we first reverse the gray levels: background as 0 and foreground in [1,255]. After that, the foreground gray levels are nonlinearly normalized to a specified range for overcoming the gray scale variation among different images [19]. For shape normalization of offline characters, we choose the line density projection interpolation (LDPI) method due to its superior performance [18]. For direction decomposition, we first compute the gradient by the Sobel operator from the original image, and then decompose the direction of gradient into its two adjacent standard chaincode directions by the parallelogram rule [60]. Note that in this process, the normalized character image is not generated, but instead, the gradient elements of original image are directly mapped to directional maps of standard image size (say, 64x64 or 32x32) incorporating pixel coordinates transformation.

*3.2. Online DirectMap*

The online HCCR datasets provide the sequences of coordinates of strokes. We also use the normalization-cooperated method for online handwritten characters, i.e., the features are extracted from the original pattern incorporating coordinate transformation without generating the normalized pattern. The shape normalization method used for online HCCR is the pseudo 2D bi-moment normalization (P2DBMN) [61], since LDPI is not applicable for online trajectory. For direction decomposition, the local stroke direction (of the line segment formed by two adjacent points) is decomposed into 8 directions and then generate the feature map of each direction [61, 21]. The imaginary strokes (pen lifts or called off-strokes) [62] are also added with a weight of 0.5 to get enhanced representation.

*3.3. Analysis*

To build compact representations, we set the size of feature map to be 32, and therefore, the generated directMap is an $8 \times 32 \times 32$ tensor. Fig. 1 shows the examples for online and offline directMaps. The first column is the original character, while the columns indexed by 0-7 are the eight directional maps. For better illustration, we also show the average map of the eight directional maps. It is shown that the shape in the average map is normalized compared with original character. For offline character, the gradient is decomposed, hence the average map gives the contour information of original image. Contrarily, for online character, the local stroke is decomposed, hence the input character can be well reconstructed by the average map, from which we can also find that the imaginary strokes are already taken into consideration. Because gradient is perpendicular to local stroke, the online and offline directMaps are different although they adopt the same direction coding as shown in the right side of Fig. 1.

DirectMap is a powerful representation for HCCR which utilizes strong prior knowledge that Chinese character is produced by basic directional strokes during writing process. As shown in Fig. 1, directMap is very sparse. Actually, in our experimental database, 92.41% (online) and 79.01% (offline) of the elements in directMap are zeros. With this sparsity, we can store and reuse the extracted directMaps efficiently. Owing to the sparsity, using maps with size smaller than the original image (more than 64x64) does not lose shape information.

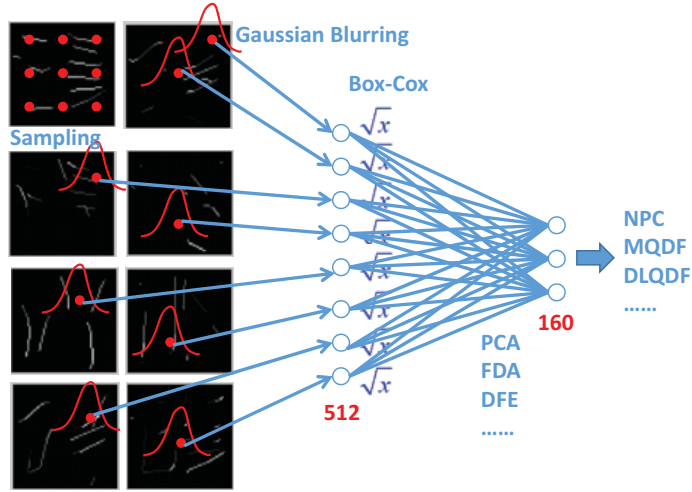

Figure 2: The traditional framework of using directMap for HCCR.

## 4. From Traditional HCCR to ConvNet

After obtaining directMaps, traditional HCCR methods [19] adopt the sampling strategy on each map. As shown in Fig. 2, at each sampling position, the Gaussian blurring [19] is used to reduce the influence of stroke position variation. Usually, $8 \times 8$ points are regularly sampled from each map, resulting in a feature vector with dimensionality 512 (for eight directions), which is widely known as the directional feature. After that, the Box-Cox transformation [63] of $y = x^{0.5}$ is applied to each feature dimension to increase the Gaussianity of data. Linear dimensionality reduction methods such as principal component analysis (PCA), Fisher discriminant analysis (FDA) [64], and discriminative feature extraction (DFE) [65] are then used to reduce the features into a low-dimensional subspace (e.g. 160). In this subspace, the nearest prototype classifier (NPC) [57], modified quadratic discriminant function (MQDF) [1], and discriminative learning quadratic discriminant function (DLQDF) [66] are widely used as final classifiers (see [67] for an overview of MQDF related methods). This kind of framework has been the benchmark for HCCR during the past decades [19, 2, 21].

Although not being clearly stated in the literature, as shown in Fig. 2, the traditional HCCR architecture is closely related to a simplified convNet. The Gaussian blurring can be viewed as a convolution mask which is pre-defined here other than learned from data. The Box-Cox transformation is a nonlinear activation although being different from the widely used activations in neural networks. After that, there is a fully-connected layer and a classification layer. Therefore, we should say that traditional HCCR methods are also following the design philosophy of deep neural networks, although the structure in Fig. 2 is really shallow and not in a standard end-to-end backpropagation training manner. In light of this, it is straightforward and necessary to integrate directMap with deep convNet to look for a new benchmark.

## 5. Convolutional Neural Network

Recently, it is shown that the depth is crucial for the success of convolutional neural networks (convNet) [24, 25]. Considering the size of our directMap ($8 \times 32 \times 32$), we build an 11-layer network for HCCR.

### 5.1. Architecture

As shown in Fig. 3, the directMap (online or offline) is passed through a stack of convolutional (conv) layers, where the filters are with a small receptive field $3 \times 3$, which is the smallest size to capture notion of left/right, up/down, and center [25]. All the convolution stride is fixed to one. The number of feature maps is increased from 50 (layer-1) to 400 (layer-8) gradually. Spatial pooling is widely used to obtain translation invariance (robustness to position). Traditionally, there is a pooling layer after each conv layer. To increase the depth of network, the size of feature map should be reduced slowly. Therefore, in our architecture, the spatial pooling is implemented after every two conv layers (Fig. 3), which is carried out by max-pooling (over a $2 \times 2$ window with stride 2) to halve the size of feature map. After the stack of 8 conv layers and 4 max-pool layers, the feature maps are flattened and concatenated into a vector with dimensionality 1600. Two fully-connected (FC) layers (with 900 and 200 hidden units respectively) are then followed. At last, the softMax layer is used to perform the 3755-way classification.

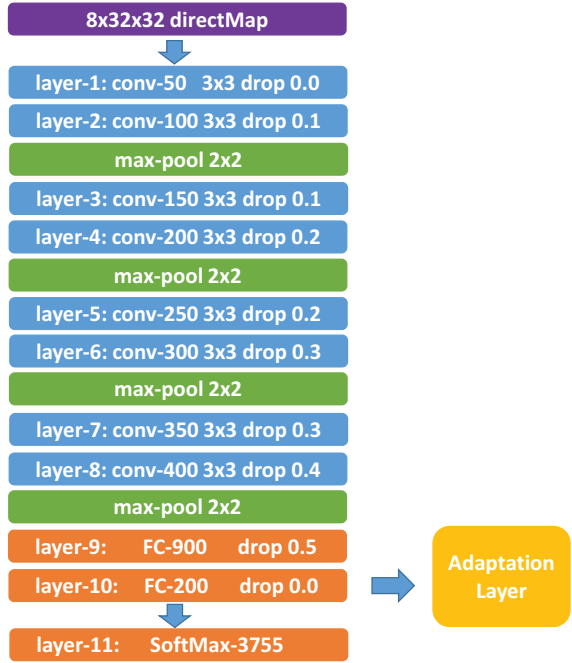

Figure 3: The convNet architecture used for both online and offline HCCR.

### 5.2. Regularization

Regularization is important for deep networks. Dropout [68] is a widely used strategy (efficient infinite model averaging) to increase generalization performance, which can be implemented by randomly dropping units (along with their connections) for each layer with a given probability. We use dropout for all the layers except layer-1 and layer-10. As shown in Fig. 3, the dropout probabilities are increased with respect to the depth. Layer-10 is the last FC layer before softMax layer, and thus can be viewed as a very high-level feature extractor. We set the dimensionality of layer-10 to be as low as 200 to obtain a compact representation (which already can be viewed as regularization), therefore, we make the dropout probability on layer-10 to be zero. Another regularization strategy we used is the weight decay with $L_2$ penalty. The multiplier for weight decay is 0.0005 during the training process.

### 5.3. Activation

Activation function is crucial for adding non-linearity into the network. Rectified linear unit (ReLU) is one of the keys to the success of deep networks [23]. A more general form named leaky-ReLU [69] is defined as $f(x) = \max(x, 0) + \lambda \min(x, 0)$ (standard ReLU use $\lambda = 0$), which can expedite convergence and obtain better performance than conventional activations (such as sigmoid and tanh). Recently, the learnable activation functions have also been used by different approaches [70, 71, 72]. By considering both the performance and efficiency, in our convNet (Fig. 3), all hidden layers are equipped with the leaky-ReLU non-linearity with $\lambda = 1/3$.

### 5.4. Initialization

The initialization of the deep network is important, because bad initializations will hamper the learning of a highly non-linear system [70]. Deep convNets are widely initialized by drawing random weights from the zero-mean Gaussian distribution $\mathcal{N}(0, \delta^2)$ or the zero-mean uniform distribution $U\left[-\sqrt{3}\delta, \sqrt{3}\delta\right]$. For example, [23] used $\mathcal{N}(0, \delta^2)$ with $\delta = 0.01$ to initialize the network. However, by using this, very deep models have difficulties to converge [25, 70]. To handle this problem, [73] and [70] proposed using different $\delta$ for different layers according to the number of input neurons (or output neurons) for a particular layer. In our experiments, we find that by using $\delta = 0.01$ for all layers, the signal shrinks as it passes through each layer and becomes too tiny when it reaches softMax layer. To ensure signal safely reaching final layer, we rescale the input data to amplify signal.

First, we initialize all the weights via $\mathcal{N}(0, \delta^2)$ ($\delta = 0.01$) and all biases as zeros. Let $s \in \mathbb{R}^c$ ($c$ is the number of classes) be the signal in the last layer before softMax activation, and $\Delta = s_{max} - s_{mean}$ be the gap between max and mean values of $s$. The softMax activation is usually implemented as $p_i = \exp(s_i - s_{max})$ and softMax$(s)_i = p_i / \sum_j p_j$. Since $\Delta$ is very small, we will get $p_{max} = 1$ and $p_{mean} \approx p_{min} \approx 1$. Hence, softMax$(s)$ results in a vector very close to $[1/c, 1/c, \ldots, 1/c]$. To handle this problem, we multiply the input data by a constant $v = -\ln 0.8/\Delta$. The purpose is to make $s$ in a suitable scale and then softMax can be effectively calculated.

Both max-pooling and leaky-ReLU have the semi-linear property of $f(vx) = vf(x)$ and the biases are initialized as zeros, therefore, the input data rescaling constant $v$ can be well preserved to the last layer. In this case, the signal in the last layer becomes $v \times s$. When we compute softMax$(v \times s)$, we will get $p_{max} = 1$ and $p_{mean} = 0.8$, which will result in a meaningful softMax calculation. In practice, the $\Delta$ is estimated as an averaged number on the training data, and $v$ (after estimation) is fixed in both training and testing.

*5.5. Training*

After proper initialization, we are able to train the network by the back-propagation algorithm [74]. The training is carried out by minimizing the multi-class negative log-likelihood loss using mini-batch gradient descent with momentum. The mini-batch size is set to be 1000, while the momentum is 0.9. The learning rate is initially set to 0.005, and then decreased by ×0.3 when the cost or accuracy on the training data stop improving. We do not use the data augmentation strategy to generate distorted samples during the training process, because we believe directMap is already a powerful representation for HCCR, and we want to make the training more efficient. The training is finished after about 70 epochs. After each epoch, we shuffle the training data to make different mini-batches. In our experiments, both the online and offline HCCR adopt the same convNet architecture (Fig. 3) and share the same training strategies as described above.

## 6. Adaptation of ConvNet

It is shown that the features extracted from the activation of a deep network trained with a large dataset can generalize to novel generic tasks [59]. However, transfer of models directly to new domains without adaptation will lead to poor performance [58]. A widely used strategy is to train or fine-tune a state-of-the-art deep model on a new domain, but this requires a significant amount of labeled data. A better solution is the domain adaptation [58, 75] of the deep models.

To adapt the deep convNet to the new handwriting style of particular writers, we propose a special adaptation layer based on our previous work [28]. The adaptation layer can be placed after any fully-connected layer in a network. Suppose $\phi(x) \in \mathbb{R}^d$ is the output (after activation) of a particular layer (let us call it source layer), and we want to put the adaptation layer after this layer. First, we estimate class-specific means on source layer from training data $\{x_i^{trn}, y_i^{trn}\}_{i=1}^N$ where $y_i^{trn} \in \{1, 2, \ldots, c\}$ and $c$ is the number of classes:

$$\mu_k = \frac{1}{\sum_{i=1}^N \mathbb{I}(y_i^{trn} = k)} \sum_{i=1}^N \phi(x_i^{trn}) \mathbb{I}(y_i^{trn} = k), \quad (1)$$

where $\mathbb{I}(\cdot) = 1$ when the condition is true and otherwise 0. The $\{\mu_1, \ldots, \mu_c\}$ represent the class distribution on source layer and will be used to learn parameters of adaptation layer.

The adaptation layer contains a weight matrix $A \in \mathbb{R}^{d \times d}$ and an offset vector $b \in \mathbb{R}^d$. There is no activation function on adaptation layer. Suppose we have some unlabeled data $\{x_i\}_{i=1}^n$ for adaptation. By passing them through the pretrained convNet, we can obtain the predictions for them $y_i = \text{convNet}_{pred}(x_i) \in \{1, \ldots, c\}$. Since the last layer of convNet is softMax, we can also get a confidence about this prediction with $f_i =$

**Algorithm 1** Unsupervised Adaptation of ConvNet
───────────────────────────────────────────
**Input:** data $\{x_i\}_{i=1}^n$, convNet, $\beta$, $\gamma$, iterNum
 1: estimate class-specific means on source layer
 2: add adaptation layer after source layer into convNet
 3: initial $A = I$ and $b = 0$
 4: **for** iter = 1 to iterNum **do**
 5:  $y_i = \text{convNet}_{\text{pred}}(x_i) \in \{1, \ldots, c\}$
 6:  $f_i = \text{convNet}_{\text{softMax}}(x_i) \in [0, 1]$
 7:  update $A$ and $b$ by (2)
 8: **end for**
**Output:** prediction $\{y_i\}_{i=1}^n$, adapted convNet
───────────────────────────────────────────

$\text{convNet}_{\text{softMax}}(x_i) \in [0, 1]$. With all these information, now the purpose of adaptation is to reduce the mismatch between the training and test data. Note that we already have the class-specific means $\mu_k$ on source layer, the adaptation problem can be formulated as:

$$\min_{A,b} \sum_{i=1}^n f_i \left\| A\phi(x_i) + b - \mu_{y_i} \right\|_2^2 + \beta \|A - I\|_F^2 + \gamma \|b\|_2^2, \qquad (2)$$

where $\|\cdot\|_F$ is the matrix Frobenius norm, $\|\cdot\|_2$ is the vector $L_2$ norm, and $I$ is the identity matrix.

The objective of adaptation as shown in (2) is to transform each point $\phi(x_i)$ towards the class-specific mean $\mu_{y_i}$ on the source layer. Since the prediction $y_i$ may be not reliable, each transformation is weighted by the confidence $f_i$ given by the softMax of the network. In practice, to guarantee the adaptation performance with small $n$, two regularization terms are adopted: the first is to constrain the deviation of $A$ from identity matrix, while the second is to constrain the deviation of $b$ from zero vector. When $\beta = \gamma = +\infty$, we will get $A = I$ and $b = 0$ which means no adaptation is happening.

After obtaining $A$ and $b$, the source layer and adaptation layer are combined together to produce an output as:

$$\text{output}(x) = A\phi(x) + b \in \mathbb{R}^d, \qquad (3)$$

which is then fed into the next layer of the network. In practice, it is better to put the adaptation layer right after the bottleneck layer in a network, i.e., the fully-connected layer which has the smallest number of hidden units compared with other layers. In this way, the size of $A$ and $b$ can be minimized, and thus the adaptation will be more efficient and effective. From this consideration, we set the dimensionality of layer-10 in our convNet (Fig. 3) to be as low as 200, and the adaptation layer is placed right after this layer.

The problem in (2) is a convex quadratic programming (QP) problem which can be solved efficiently with a closed-form solution [28]. We use a self-training strategy for unsupervised adaptation of convNet. As shown in Algorithm 1, we first initialize the adaptation as $A = I$ and $b = 0$. After estimating $y_i$ and $f_i$ from convNet, we update $A$ and $b$ according to (2). With new parameters of $A$ and $b$, the network prediction $y_i$ and softMax confidence $f_i$ will be more accurate. Therefore, we repeat this process several times to automatically boost the performance (see Algorithm 1). It is straightforward to use our method for supervised adaptation by discarding the self-training iteration and using the labels provided by data. However, unsupervised adaptation is more common in practice, since unlabeled data are much easier to obtain, and unsupervised adaptation can apply to test data directly.

Adding an adaptation layer in convNet will not change the structures and parameters of other layers. If we want to return to an unadapted situation, we can simply set $A = I$ and $b = 0$. The adaptation process is an efficient QP which do not require sophisticated optimization tricks. With the involved regularization, the adaptation performance is guaranteed, and the adaptation data are not required to cover all the classes. Moreover, the adaptation layer can be efficiently implemented in an unsupervised manner. These are the advantages of our method compared with the fine-tuning based strategies.

| Offline: | #writers | #samples fall in 3755-class | ratio |
|---|---|---|---|
| HWDB1.0 | 420 | 1,556,675 (train) | 58.02% |
| HWDB1.1 | 300 | 1,121,749 (train) | 41.81% |
| HWDB1.2 | 300 | 4,463 (train) | 00.17% |
| Off-ICDAR2013 | 60 | 224,419 (test) | – |
| Online: | #writers | #samples fall in 3755-class | ratio |
| OLHWDB1.0 | 420 | 1,570,051 (train) | 58.20% |
| OLHWDB1.1 | 300 | 1,123,132 (train) | 41.63% |
| OLHWDB1.2 | 300 | 4,490 (train) | 00.17% |
| On-ICDAR2013 | 60 | 224,590 (test) | – |

Table 1: The statistics for online and offline HCCR databases.

## 7. Experiments

We conduct experiments for both online and offline HCCR to compare our methods with other previously reported state-of-the-art approaches. After that, the effectiveness for the adaptation of convNet is evaluated on 60 writers from the ICDAR-2013 competition database.

### 7.1. Database

For training the convNets, the databases collected by CASIA [76] can be used as training sets, which contain the offline handwritten character datasets HWDB1.0-1.2 and the online handwritten character datasets OLHWDB1.0-1.2. In the following sections, we denote these datasets (either offline or online) simply as DB1.0-1.2. The test data are the ICDAR-2013 offline and online competition datasets [10] respectively, which were produced by 60 writers different from the training datasets. The number of character classes is 3755 (level-1 set of GB2312-80). Table 1 shows the statistics of all the datasets. Note that the size of DB1.2 is negligible compared with DB1.0 and DB1.1, this is because most of the characters in DB1.2 are out of the vocabulary of the 3755-class considered in the ICDAR-2013 competition.

### 7.2. Training of ConvNet

We train convNets on the directMaps for HCCR. The details on the generating of directMaps are described in Section 3. Two convNets (with the same architecture in Fig. 3) are trained for online and offline HCCR separately, due to the difference in their directMaps (Fig. 1). We use the method of SGD with momentum for training other than the adaptive learning rate methods such as RMSProp, AdaGrad, AdaDelta [77], and Adam [78]. The learning rate is dropped by ×0.3 when the cost or accuracy on training data is not improving. In practice, we find this learning rate decay strategy (also used by [23, 70, 25]) is very effective. As shown in Fig. 4, initially the training is getting improved smoothly and then it sink into a plateaus, but after the first learning rate drop, there is a significant accuracy improvement for both training and test data. The learning rate is initialized as 0.005 and reduced three times prior to termination. Other parameters involved in training the convNet can be found in Section 5.5. Our convNet is trained with the Theano [79, 80] platform using a NVIDIA Titan-X 12G GPU. Training one network takes about 80 hours to converge, and the total number of epochs is about 70 for each model.

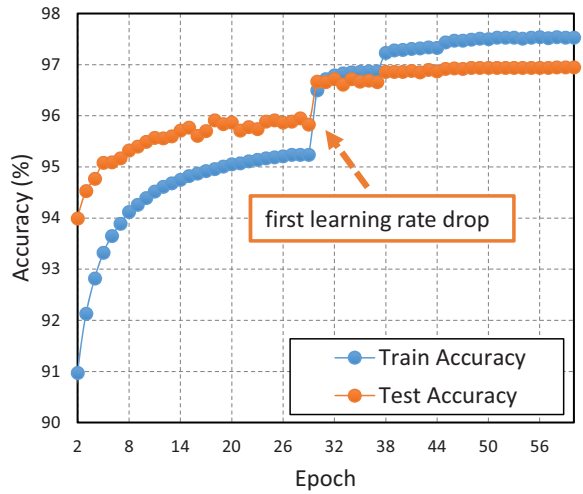

Figure 4: The training process of the convNet for offline HCCR.

| Results on ICDAR-2013 Offline HCCR Competition Database | | | | | | | |
|---|---|---|---|---|---|---|---|
| No. | Method | Ref. | Accuracy | Memory | Training Data | Distortion | Ensemble |
| 1 | Human Performance | [10] | 96.13% | n/a | n/a | n/a | n/a |
| 2 | Traditional Method: DFE + DLQDF | [19] | 92.72% | 120.0MB | 1.0+1.1 | no | no |
| 3 | CCPR-2010 Winner: HKU | [8] | 89.99% | 339.1MB | 1.0+1.1 | yes | no |
| 4 | ICDAR-2011 Winner: IDSIAnn-2 | [9] | 92.18% | 27.35MB | 1.1 | yes | no |
| 5 | ICDAR-2013 Winner: Fujitsu | [10] | 94.77% | 2.402GB | 1.1 | yes | yes (4) |
| 6 | Multi-Column DNN (MCDNN) | [38] | 95.79% | 349.0MB | 1.1 | yes | yes (8) |
| 7 | ATR-CNN Voting | [40] | 96.06% | 206.5MB | 1.1 | yes | yes (4) |
| 8 | HCCR-Gabor-GoogLeNet | [15] | 96.35% | 27.77MB | 1.0+1.1 | no | no |
| 9 | HCCR-Ensemble-GoogLeNet-4 | [15] | 96.64% | 110.9MB | 1.0+1.1 | no | yes (4) |
| 10 | HCCR-Ensemble-GoogLeNet-10 | [15] | 96.74% | 270.0MB | 1.0+1.1 | no | yes (10) |
| 11 | CNN-Single | [41] | 96.58% | 190.0MB | 1.0+1.1+1.2 | yes | no |
| 12 | CNN-Voting-5 | [41] | 96.79% | 950.0MB | 1.0+1.1+1.2 | yes | yes (5) |
| 13 | DirectMap + ConvNet | ours | **96.95%** | 23.50MB | 1.0+1.1 | no | no |
| 14 | DirectMap + ConvNet + Adaptation | ours | **97.37%** | 23.50MB | 1.0+1.1 | no | no |
| 15 | DirectMap + ConvNet + Ensemble-2 | ours | 97.07% | 47.00MB | 1.0+1.1 | no | yes (2) |
| 16 | DirectMap + ConvNet + Ensemble-3 | ours | 97.12% | 70.50MB | 1.0+1.1 | no | yes (3) |

Table 2: Different methods for ICDAR-2013 offline HCCR competition.

*7.3. Offline HCCR Results*

Table 2 shows the results of different methods on ICDAR-2013 offline competition database. From Table 2, we can find that there is a large gap between the traditional method (2nd row) and the human-level performance (1st row). Through the three competitions, the recognition accuracies are gradually increased, which identify the effectiveness of holding competition for promoting researches. In ICDAR-2011, the team from IDSIA of Switzerland (4th row) won the first place [9], and in ICDAR-2013, the team from Fujitsu (5th row) took the first place [10]. After correcting a bug in their system [38], the team of IDSIA again achieved the best performance (6th row). After that, by improving their method [40], the team from Fujitsu boosted their performance as shown in 7th row. The human-level performance was firstly surpassed by [15] (8th row) with their Gabor-GoogLeNet. By using the ensemble of ten models, their accuracy was further improved to 96.74% (10th row). Recently, the team from Fujitsu [41] further improved their system by using proper sample generation (local and global distortion), multi-supervised training, and multi-model ensemble. They achieved 96.58% by a single network (11th row), and with the ensemble of 5 networks, their accuracy was improved to 96.79% (12th row), which is the best reported result for offline HCCR so far.

Our proposed method of directMap+convNet can achieve a new benchmark for offline HCCR as shown in the 13th row of Table 2. Particularly, our result is based on a single network. Compared with the best single models (8th and 11th rows in Table 2), our model achieves significant accuracy improvement. Moreover, our method also has the lowest memory usage compared with all the other systems, due to the compact representation of directMap and our special convNet structure. For example, the previous best performance achieved by [41] (12th row) is based on the ensemble of 5 networks with memory consumption 950MB, while our model (13th row) is a single network of 23.5MB. In Table 2, the methods from 4th to 16th are all based on convNets, which implies that deep learning based methods are becoming more and more popular for solving offline HCCR problem. However, the domain-specific knowledge is not used by 4th-7th methods, and hence their accuracies are much lower than ours. Although domain-specific knowledge is already used by 8th-10th methods, our approach (with a single network) can still outperforms them. The 11th and 12th methods used multiple strategies to improve their performance such as very deep (15-layer) network, proper designed distortion methods, and multi-supervised training strategy [81]. However, our method still outperforms them by using only a single objective function in the training process and without data augmentation. These results clearly identify the advantages of integrating directMap with convNet used in our approach for HCCR.

| | Results on ICDAR-2013 Online HCCR Competition Database | | | | | | |
|---|---|---|---|---|---|---|---|
| No. | Method | Ref. | Accuracy | Memory | Training Data | Distortion | Ensemble |
| 1 | Human Performance | [10] | 95.19% | n/a | n/a | n/a | n/a |
| 2 | Traditional Method: DFE + DLQDF | [19] | 95.31% | 120.0MB | 1.0+1.1 | no | no |
| 3 | CCPR-2010 Winner: SCUT-HCII-2 | [8] | 92.39% | 30.06MB | 1.0+1.1 | no | yes (2) |
| 4 | ICDAR-2011 Winner: VO-3 | [9] | 95.77% | 41.62MB | 1.0+1.1+Others | yes | no |
| 5 | ICDAR-2013 Winner: UWarwick | [10] | 97.39% | 37.80MB | 1.0+1.1+1.2 | yes | no |
| 6 | DropSample-DCNN | [43] | 97.23% | 15.00MB | 1.0+1.1 | yes | no |
| 7 | DropSample-DCNN-Ensemble | [43] | 97.51% | 135.0MB | 1.0+1.1 | yes | yes (9) |
| 8 | DirectMap + ConvNet | ours | **97.55%** | 23.50MB | 1.0+1.1 | no | no |
| 9 | DirectMap + ConvNet + Adaptation | ours | **97.91%** | 23.50MB | 1.0+1.1 | no | no |
| 10 | DirectMap + ConvNet + Ensemble-2 | ours | 97.60% | 47.00MB | 1.0+1.1 | no | yes (2) |
| 11 | DirectMap + ConvNet + Ensemble-3 | ours | 97.64% | 70.50MB | 1.0+1.1 | no | yes (3) |

Table 3: Different methods for ICDAR-2013 online HCCR competition.

*7.4. Online HCCR Results*

The comparison of different methods on ICDAR-2013 online competition database is shown in Table 3. For online HCCR, the traditional method (2nd row) is already better than human performance (1st row), and the human performance for online HCCR is much lower than offline HCCR. This is because the display of online characters (stroke coordinates) as still images is not as pleasing as that of offline samples. The three competitions also exhibit evident progresses for online HCCR. The best single network (5th row) is the ICDAR-2013 winner from University of Warwick, which represents the characteristics of stroke trajectory with a "signature" from the theory of differential equations [12] and adopts a sparse structure of convolutional neural network [39]. Recently, by combining multiple domain knowledge and using a "dropSample" training strategy, [43] can achieve comparable performance by using a single network (6th row). With the ensemble of nine networks, they further improve the performance to 97.51% (7th row), which is the best reported result for online HCCR so far.

With our directMap+convNet, we can set a new benchmark for online HCCR as shown in the 8th row of Table 3. Our result is based on a single network. Compared with the best single network method from the 5th row, our accuracy and memory usage are both better. Although the ensemble model (7th row) has a comparable accuracy with us, our memory usage is much more efficient, because they use the ensemble of 9 models to reach the performance. Furthermore, in order to achieve their best results, all the 4th to 7th methods adopt the data augmentation strategy with distorted samples to enlarge their training set. Contrarily, our result is achieved without data augmentation, which makes the training process to be more efficient. These comparisons again verify the effectiveness of directMap+convNet for HCCR.

*7.5. Discussion*

With directMap+convNet, we can achieve new benchmarks for both online and offline HCCR without data augmentation and model ensemble. From Table 2 and Table 3, we can find that previous best results on online and offline HCCR are achieved by different methods which have different network architectures. However, with our method, we can obtain best results for both online and offline HCCR under the same framework. We also show the top-N accuracies of our methods in Fig. 5. It is shown that the top-2 accuracies for offline and online HCCR are both higher than 99%. This will significantly benefits the handwritten Chinese text recognition [5, 6] for which the language model can be used to further improve the initial prediction, and hence, high top-N accuracy is crucial.

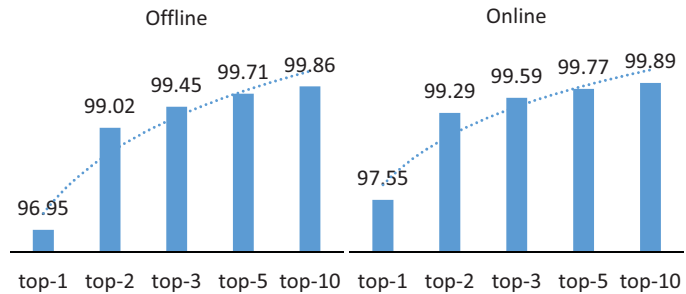

Figure 5: The top-N accuracies (%) for online and offline HCCR.

In practice, another important issue is the prediction speed of the system. It is impossible to compare different methods under the same software and hardware configuration, because this needs to re-implement so many models. We report the prediction speed in Table 4 to ease the future comparison with our system. Our system is a combination of directMap and convNet. It is shown that the extraction of directMap is very efficient although being implemented on CPU. Moreover, GPU is very important for the speedup of convNet. Compared with the ICDAR-2013 winners, our methods are faster for both online and offline HCCR. The memory usage of our system is very efficient (see Tables 2 and 3) which allow it to be used in different environments such as handheld mobile devices. However, the running time of convNet without GPU is still a shortcoming which prevents convNet from wide applications. Therefore, speedup of convNet (in non-GPU situation) is an important future direction.

*7.6. Varying the Number of Training Data*

HCCR is a large category problem (tens of thousands of different classes), and in the competition of ICDAR-2013, a standard set of 3755-class is considered. All participants were suggested to use the datasets of DB1.0-1.2 as shown in Table 1 for training their system. The test data were kept confidential until the end of competition. The participants were allowed to use external data to enlarge their training set. Nevertheless, as shown in Tables 2 and 3, all the methods were using the recommended training sets, except the 4th method in Table 3 which was submitted by a company and some private data were used by them. Moreover, as shown in Table 1, the size of DB1.2 is negligible, and therefore, the system trained with DB1.0+1.1 can be fairly compared with the system trained with DB1.0+1.1+1.2.

Although most methods can be compared fairly, we found that there are some methods (i.e., 4th to 7th in Table 2) using significant less training data compared with other approaches. To make a fair comparison with these methods, we re-trained our convNet with only HWDB1.1, and the test accuracy became 96.55%. Note that this is achieved with a single network without data augmentation. The best performance (trained with only HWDB1.1) is from the 7th method in Table 2, which was produced by the ensemble of 4 networks to give the accuracy of 96.06%. Note that our performance is even better than the 8th method in Table 2 although they used HWDB1.0+1.1 (twice more) as training data. Moreover, our performance is comparable with the 11th method in Table 2 which used all the DB1.0+1.1+1.2 (with distortion) as training set. These results again verify the effectiveness of our approach in case of smaller number of training data.

|  | Offline HCCR | Online HCCR |
|---:|---:|---:|
| directMap (CPU) | 1.9970 ms | 0.4654 ms |
| convNet (CPU) | 296.8941 ms | 294.5713 ms |
| convNet (GPU) | 0.4641 ms | 0.4556 ms |
| ICDAR-2013 Winners | GPU: 55 ms | CPU: 355 ms |

Table 4: Processing time for one character (in millisecond).

|  | Ref. | Representation Size | Type |
|---|---|---|---|
| MCDNN | [38] | $1 \times 48 \times 48$ | offline |
| ATR-CNN | [40] | $1 \times 48 \times 48$ | offline |
| GoogLeNet | [15] | $17 \times 120 \times 120$ | offline |
| SparseConvNet | [12] | $7 \times 96 \times 96$ | online |
| DropSample | [43] | $30 \times 96 \times 96$ | online |
| DirectMap | ours | $8 \times 32 \times 32$ | both |

Table 5: Size of different representations.

|  | Test Accuracy |
|---|---|
| No Gray-Level Normalization | 96.93% |
| Linear Gray-Level Normalization | 96.90% |
| Nonlinear Gray-Level Normalization | 96.95% |

Table 6: Comparison of different preprocessing methods.

*7.7. Comparison of Different Representations*

The representation of handwritten characters is an important issue. Since offline characters are naturally stored as scanned images, many approaches directly train their convNet on the raw data such as the 4th-7th methods in Table 2. Recently, it is shown by [41] that very high performance can be obtained by directly training on raw data. However, deeper network, proper sample distortion, and multi-supervised training [81] are required to guarantee the performance. It is shown in [41] that: without the multi-supervised training, the network is not able to converge during the training. Moreover, to eliminate the influence from different handwriting styles, convNet trained on raw data is usually designed to be very large (e.g. 190MB for the 11th method in Table 2) and distortion is usually required to enlarge the training set. Contrarily, by integrating domain-specific knowledge into the representation, the within-class variation can be reduced, then the convNet can be much smaller and data distortion is no longer needed (see 8th and 13th methods in Table 2). For online handwritten characters, to make full use of the temporal and spatial information, the path signature feature map was used by [12] to win the ICDAR-2013 online competition. Recently, [43] further combined path signature feature with other domain-specific knowledge (resulting in 6 different domain knowledge layers) to enhance the representation for online HCCR.

We compare the size of different representations in Table 5. It is shown that our directMap is a compact representation compared with other approaches. For example, in offline HCCR, the method of [15] used 17 feature maps ($120 \times 120$) including eight Gabor maps, eight gradient maps, and one HoG map. Furthermore, in online HCCR, as many as 30 feature maps ($96 \times 96$) were used by [43] to achieve high performance. Compared with these approaches, our method adopt the $8 \times 32 \times 32$ directMap as input to convNet, therefore, both the training and testing are much more efficient. Larger map size can also be used for our directMap, however, this requires much more computational resource. Moreover, as shown in Section 3.3, the directMap is already very sparse, therefore, enlarging the map size will not bring much information for distinguishing different classes. The highest accuracies and compact representations of our approach verify the advantages of using directMap for representing handwritten characters.

*7.8. Effectiveness of Gray-Level Normalization*

The preprocessing step is important for the success of a pattern recognition system. For example, in [38] the preprocessing includes: rescaling the images to a fixed size and contrast maximization. However, due to a glitch in their system, i.e., in training process Matlab was used while in testing process OpenCV was used, they lost the first place in ICDAR-2013 offline competition [10]. By correcting this bug, their performance can be significantly improved [38], which can outperform the ICDAR-2013 winner.

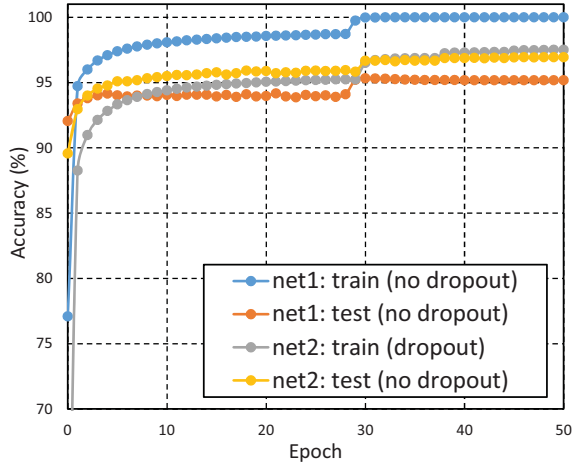

Figure 6: Comparison of two networks trained with and without dropout.

In our system, as described in Section 3.1, there also exists a preprocessing step of gray-level normalization. To check the influence of this step for the whole system, we conducted three experiments: no gray-level normalization, linear gray-level normalization, and nonlinear gray-level normalization [19]. The comparison is shown in Table 6. It is shown that the nonlinear normalization is slightly better than other methods, but the advantage is not significant. In traditional approach, each category (class) is assumed to be a Gaussian distribution [1], and therefore, nonlinear gray-level normalization is important for satisfying this assumption and then improving the final classification accuracy [19]. However, for deep learning based system, there is no significant difference between different methods as shown in Table 6. Therefore, we can conclude that this preprocessing step is not crucial for our deep learning based system. However, to avoid the mistake reported in [38], we should keep in mind that identical preprocessing should be applied for both training and testing.

### 7.9. Effectiveness of Dropout

Dropout [68] is a widely used strategy to avoid overfitting and improve the generalization performance. Previous best approaches on either offline [41] or online [12, 43] HCCR have adopted dropout to improve the performance of their deep system. In our network architecture as shown in Fig. 3, the dropout probability is increased with respect to the depth of the network. The purpose of doing so is that the bottom layers of a deep network are usually harder to train compared with the top layers, and therefore, we set smaller dropout probabilities for the bottom layers. We conduct experiments on the offline HCCR to show the effectiveness of dropout. As shown in Fig. 6, without dropout (net1), the training accuracy is increased very fast, and after 30 epochs, it reaches 100%. However, the test accuracy of net1 is not promising due to the overfitting on training data. This indicates that without dropout, in order to avoid overfitting, smaller network size should be used to reduce the capacity of the network. However, the performance of a small network (without dropout) is not as good as a big network (with dropout), because the latter can be viewed as the ensemble of many sub-networks [68] (although only one network is trained). As shown in Fig. 6, with the help of dropout (net2), the network can be effectively trained during many epochs and the training accuracy is still controlled (not growing too fast) to avoid overfitting. Note that in testing process, dropout is no longer used for both net1 and net2, and the test accuracy of net2 is significantly better than net1. This proves the effectiveness of dropout in improving the generalization performance of deep neural network.

### 7.10. Adaptation Results

As shown in Fig. 7, the same character can be written in different handwriting styles by different individuals. This is another major challenge for HCCR. Compared with traditional methods and human-level performance, deep learning based methods can achieve much higher accuracies. Nevertheless, in this section, we will show that writer adaptation of deep convNet is still effective in further improving the performance. The ICDAR-2013 competition database [10] contains 60 writers. Each writer was supposed to produce 3755 characters (one for each class), but some of the characters were eliminated due to miss-labeling or low-quality. To make a fair comparison with other methods, the labels provided by the competition database should only be used for accuracy evaluation. Therefore, we consider the unsupervised adaptation of convNet as described in Section 6. In other words, the only additional information we used is the writer index in the database. Traditional evaluation treats the test data to be independent, but here, the data are partitioned into 60 groups according to the writer index. Each group of data is produced by one writer and assumed to be consistent in writing style. The data of one writer are then fed into Algorithm 1 to obtain the final prediction.

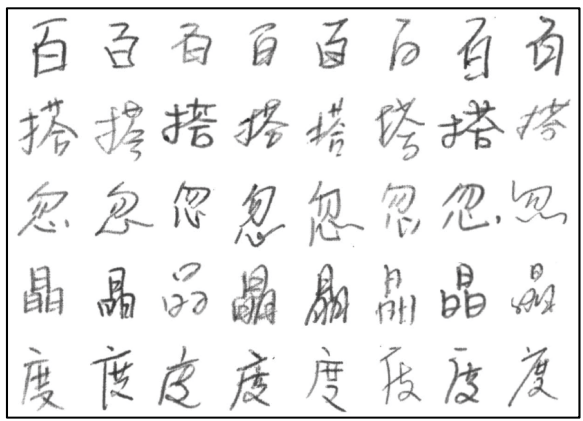

Figure 7: Different handwriting styles: each row represents the same character written by different people.

The hyper-parameters of Algorithm 1 include the trade-off parameter $\beta, \gamma$ and the number of iterations for self-training. To achieve better performance, $\beta$ is set to be $\widetilde{\beta}$ multiplied by a constant which is estimated from data (see eq. (10) in [28] for more detail). After that, for all the 60 writers, we set $\widetilde{\beta} = 0.2$ and $\gamma = 0$ as suggested by [28]. The iterNum in Algorithm 1 is set to be 3, because in practice only a small number of self-training is good enough to obtain stable results.

To analyze the behavior of adaptation on individuals, we consider the error reduction rate which is defined as:

$$\text{Error reduction rate} = \frac{\text{Error}_{\text{initial}} - \text{Error}_{\text{adapted}}}{\text{Error}_{\text{initial}}}. \quad (4)$$

The plots of 60 writers by their error reduction rate and initial accuracy are shown in Fig. 8 (offline) and Fig. 9 (online). We can find that: all the error reduction rates are larger than zero (except one writer), which means after adaptation the accuracies are consistently improved for both online and offline HCCR. It is also revealed that: for the writers with high initial accuracy, more improvements can be obtained with adaptation. This is because the success of self-training relies on the initial prediction. For example, in the top-right corner of Fig. 8, there is a writer whose initial accuracy is already as high as 99.33%, but after adaptation it is improved to 99.63%, given an error reduction rate of more than 44%.

The average accuracies of the 60 writers after adaptation are shown in the 14th row of Table 2 and 9th row of Table 3 respectively. For offline HCCR, the accuracy is improved from 96.95% to 97.37%, while for online HCCR, the accuracy is improved from 97.55% to 97.91%. Note that in the adaptation process, we only add an adaptation layer into the network with parameters $A \in \mathbb{R}^{200 \times 200}$ and $b \in \mathbb{R}^{200}$, which are negligible compared with the full size of convNet. The number of writer-specific data used for adaptation is equal to (or less than) the number of classes. Moreover, the whole process is happened in an unsupervised manner. Consider all these together, we can conclude that the proposed adaptation layer is effective in improving the accuracy of convNet. Furthermore, the adaptation is also very efficient, because only a QP problem with closed-form solution is involved for each iteration and three iterations are used for each adaptation (each writer).

*7.11. Model Ensemble Results*

The ensemble of multiple models is widely used to achieve the highest accuracies on different databases [37, 15, 43, 41]. Model ensemble can be achieved by training the same convNet multiple times with different random initializations and different mini-batches (due to random data shuffling). We train our convNet (Fig. 3) three times for both online and offline HCCR. To make the final prediction, the softMax probabilities of different models are averaged for classification. The results of model ensemble are shown in the 15th-16th rows of Table 2 and 10th-11th rows of Table 3 respectively. It is shown that model ensemble can indeed improve the classification accuracies. We did not consider the ensemble of more than three models, because the performance gain from two models to three models is already vanishing, and the usefulness of model ensemble is limited in real applications. The ensemble of two models already double the processing time and memory usage. With more models involved, more time and space resources will be consumed. By comparing the improvements caused by model ensemble and adaptation, we can conclude that the adaptation of convNet is much more efficient and effective. Therefore, in practice, it is much better to use a single network with adaptation for high accuracy personalized handwriting recognition, other than the ensemble of many networks.

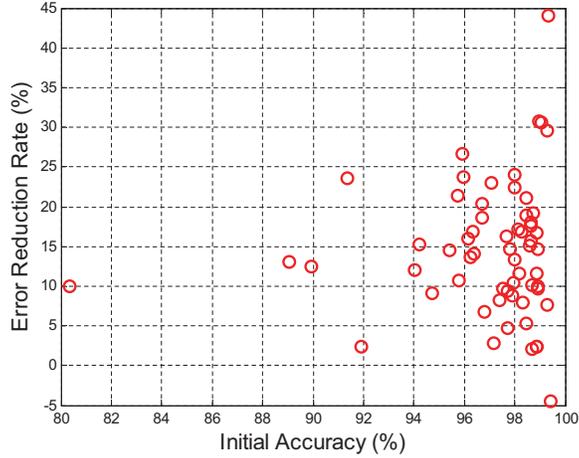

Figure 8: The error reduction rates of 60 writers for offline HCCR.

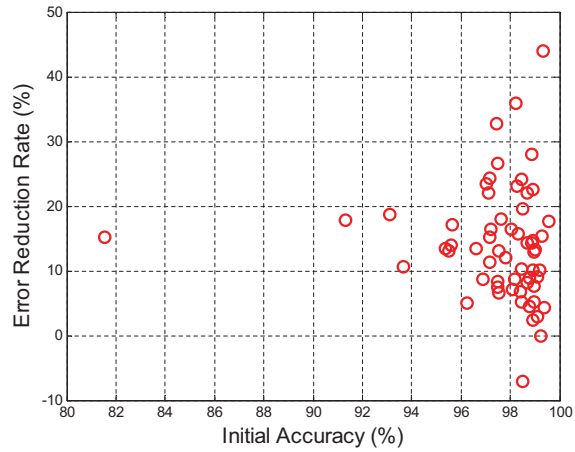

Figure 9: The error reduction rates of 60 writers for online HCCR.

*7.12. Shape Normalization and Direction Decomposition*

The proposed directMap utilizes strong prior knowledge of shape normalization and direction decomposition. As discussed in Section 3, the relationship between shape normalization and direction decomposition can be interpreted in two different ways: normalization-based and normalization-cooperated [22]. For normalization-based method, direction decomposition is performed on the normalized character. Contrarily, for normalization-cooperated method, direction decomposition is implemented on original character, and shape normalization only serves as a coordinate mapping for generating the directional maps [22].

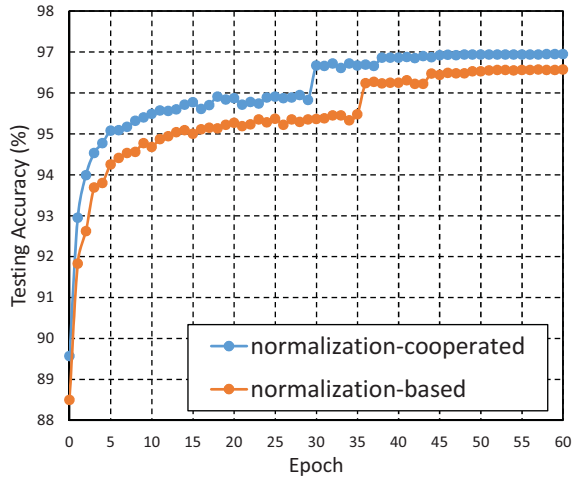

Figure 10: The comparison of normalization-cooperated and normalization-based directMaps on training the convNet for offline HCCR.

We compare the normalization-based and normalization-cooperated directMaps in training the convNet for offline HCCR. As shown in Fig. 10, the normalization-cooperated directMap significantly outperform the normalization-based one. This is because: although shape normalization can reduce the within-class variance, it also causes stroke shape distortion. Normalization-cooperated method can well preserve the original stroke direction and also benefit from shape normalization, therefore it is much more discriminative than normalization-based method. Using state-of-the-art shape normalization methods (i.e., LDPI for offline and P2DBMN for online HCCR) and normalization-cooperated direction decomposition is the key for the success of our directMaps, which can give the highest recognition accuracies when combined with deep convNets as shown in the previous sections.

## 8. Conclusion

By integrating the deep convolutional neural network (convNet) with the domain-specific knowledge of shape normalization and direction decomposition (directMap), this paper sets new benchmarks for both online and offline HCCR on the ICDAR-2013 competition database. The directMap is shown to be a compact and powerful representation for handwritten characters. Combining directMap with an 11-layer convNet can achieve highest accuracies without the help from data augmentation and model ensemble. Compared with previous state-of-the-art methods, our approach is much better from the aspects of both recognition accuracy and memory usage. Since high character recognition accuracy is essential for the success of handwritten text recognition [5, 6], our future work is to extend the proposed methods and set new benchmarks for handwritten Chinese text recognition.

Although deep learning based methods can outperform the traditional approaches with large margins, this paper shows that writer adaptation of deep networks can still improve the performance consistently and significantly. The adaptation is achieved by adding an efficient and effective adaptation layer into the network. The objective of the adaptation is to reduce the mismatch between training and test data on the source layer, which is formulated into an efficient QP problem and can be solved with closed-form solution. The adaptation performance is guaranteed even with a small amount of adaptation data. Moreover, the whole adaptation process can be effectively implemented in an unsupervised manner. Since convNet has been successfully used for many other problems, it is straightforward and interesting to evaluate the adaptation layer on other visual recognition tasks such as image classification, object detection, face recognition and so on. Besides convNet, our future work will also consider using or extending the adaptation layer for other models such as the recurrent neural network [82].

Recent works of HCCR using deep learning methods have surpassed the human performance. However, this is not the end of research. Further improvement and speedup of convNet can be achieved with better network architecture (such as the super deep network [83]) or training algorithm. Moreover, the high accuracy of a handwritten character recognition system will enable and inspire many other related tasks. In future, the three basic components (directMap, convNet, and adaptation) can be hopefully combined with other models (such as recurrent neural network) to solve other challenging problems such as handwritten Chinese text recognition, handwritten document analysis and retrieval, natural scene (video) text detection and recognition, historical document analysis, and so on.

## Acknowledgments

The authors thank the developers of Theano [79, 80] for providing such a good deep learning platform.